\documentclass[sigconf]{acmart}
\AtBeginDocument{%
  }

\copyrightyear{2025}
\acmYear{2025}
\setcopyright{acmlicensed}\acmConference[MM '25]{Proceedings of the 33rd ACM International Conference on Multimedia}{October 27--31, 2025}{Dublin, Ireland}
\acmBooktitle{Proceedings of the 33rd ACM International Conference on Multimedia (MM '25), October 27--31, 2025, Dublin, Ireland}
\acmDOI{10.1145/3746027.3754879}
\acmISBN{979-8-4007-2035-2/2025/10}

\usepackage{multicol}
\usepackage{multirow}
\usepackage{utfsym}
\usepackage{tablefootnote}
\usepackage{pifont}
\usepackage{colortbl}
\usepackage{balance}

\begin{document}

\title{Serial Over Parallel: Learning Continual Unification for Multi-Modal Visual Object Tracking and Benchmarking}

\author{Zhangyong Tang}
\email{zhangyong_tang_jnu@163.com}
\affiliation{%
  \institution{Jiangnan University}
  \city{Wuxi}
  \country{China}}
\orcid{0001-8187-9384}

\author{Tianyang Xu}
\email{tianyang.xu@jiangnan.edu.cn}
\affiliation{%
  \institution{Jiangnan University}
  \city{Wuxi}
  \country{China}}
\orcid{0002-9015-3128}

\author{Xue-Feng Zhu}
\email{xuefeng_zhu95@163.com}
\affiliation{%
  \institution{Jiangnan University}
  \city{Wuxi}
  \country{China}}
\orcid{0003-0262-5891}

\author{Chunyang Cheng}
\email{chunyang_cheng@163.com}
\affiliation{%
  \institution{Jiangnan University}
  \city{Wuxi}
  \country{China}}
\orcid{0003-4603-3505}

\author{Tao Zhou}
\email{taozhou.ai@gmail.com}
\affiliation{%
 \institution{Nanjing University of Science and Technology}
 \city{Nanjing}
 \country{China}}
\orcid{0002-3733-7286}

\author{Xiaojun Wu*}
\email{wu_xiaojun@jiangnan.edu.cn}
\affiliation{%
  \institution{Jiangnan University}
  \city{Wuxi}
  \country{China}}
\orcid{0000-0199-5001}

\author{Josef Kittler}
\email{j.kittler@surrey.ac.uk}
\affiliation{%
  \institution{University of Surrey}
  \city{Guildford}
  \country{United Kingdom}}
\orcid{0002-8110-9205 }

\renewcommand{\shortauthors}{Zhangyong Tang et al.}

\begin{abstract}
Unifying multiple multi-modal visual object tracking (MMVOT) tasks draws increasing attention due to the complementary nature of different modalities in building robust tracking systems.
Existing practices mix all data sensor types in a single training procedure, structuring a parallel paradigm from the data-centric perspective and aiming for a global optimum on the joint distribution of the involved tasks. 
However, the absence of a unified benchmark where all types of data coexist forces evaluations on separated benchmarks, causing \textit{inconsistency} between training and testing, thus leading to performance \textit{degradation}.
To address these issues, this work advances in two aspects: \ding{182} A unified benchmark, coined as UniBench300, is introduced to bridge the inconsistency by incorporating multiple task data, reducing inference passes from three to one and cutting time consumption by 27\%.
\ding{183} The unification process is reformulated in a serial format, progressively integrating new tasks.
In this way, the performance degradation can be specified as knowledge forgetting of previous tasks, which naturally aligns with the philosophy of continual learning (CL), motivating further exploration of injecting CL into the unification process.
Extensive experiments conducted on two baselines and four benchmarks demonstrate the significance of UniBench300 and the superiority of CL in supporting a stable unification process.
Moreover, while conducting dedicated analyses, the performance degradation is found to be negatively correlated with network capacity.
Additionally, modality discrepancies contribute to varying degradation levels across tasks (RGBT > RGBD > RGBE in MMVOT), offering valuable insights for future multi-modal vision research. 
Source codes and the proposed benchmark is available at \textit{https://github.com/Zhangyong-Tang/UniBench300}.
\end{abstract}


\begin{teaserfigure}
  \includegraphics[width=\textwidth]{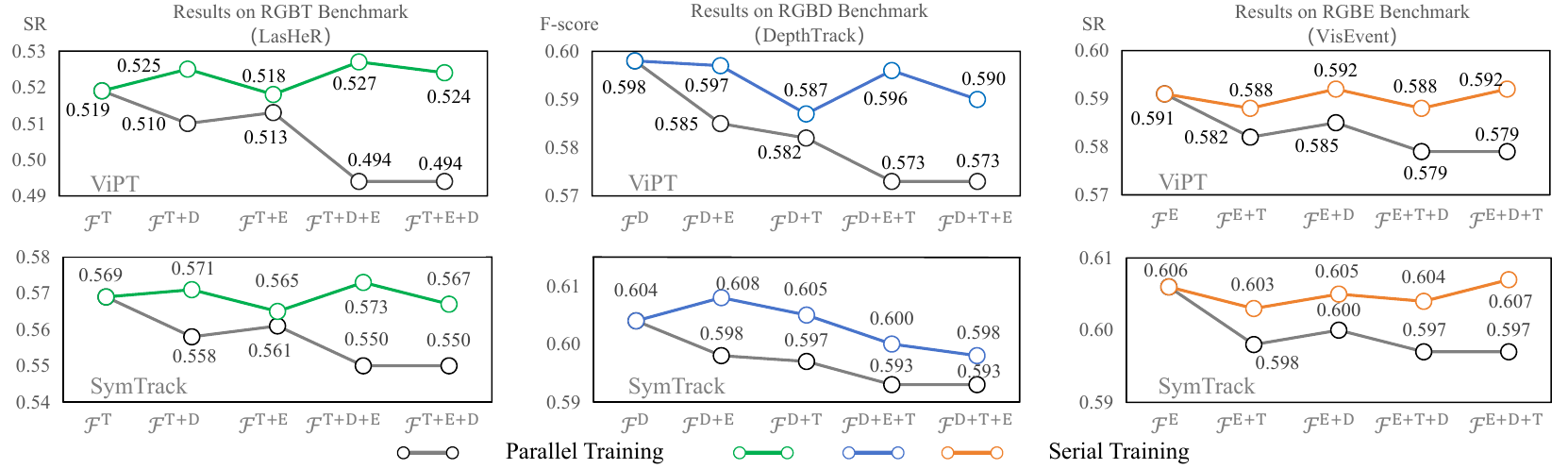}
  \caption{Comparisons between unified trackers trained in serial and parallel manners. $\rm \mathcal{F} ^{T}$, $\rm \mathcal{F}^{D}$, and $\rm \mathcal{F}^{E}$ represent tasks with RGBT (visible (RGB)+thermal infrared (T)), RGBD (RGB+depth (D)), and RGBE (RGB+event (E)) data, respectively. $\rm \mathcal{F}^{T+D}$ indicates a scenario where RGBT data is previously available and RGBD data is subsequently introduced. In this scenario, RGBT and RGBD data are mixed disorderly in parallel training while sequentially incorporated in serial training.
  }
  \label{fig:teaser}
\end{teaserfigure}


\maketitle

\section{Introduction}


\begin{figure*}[t]
  \centering
  \includegraphics[width=1.0\linewidth]{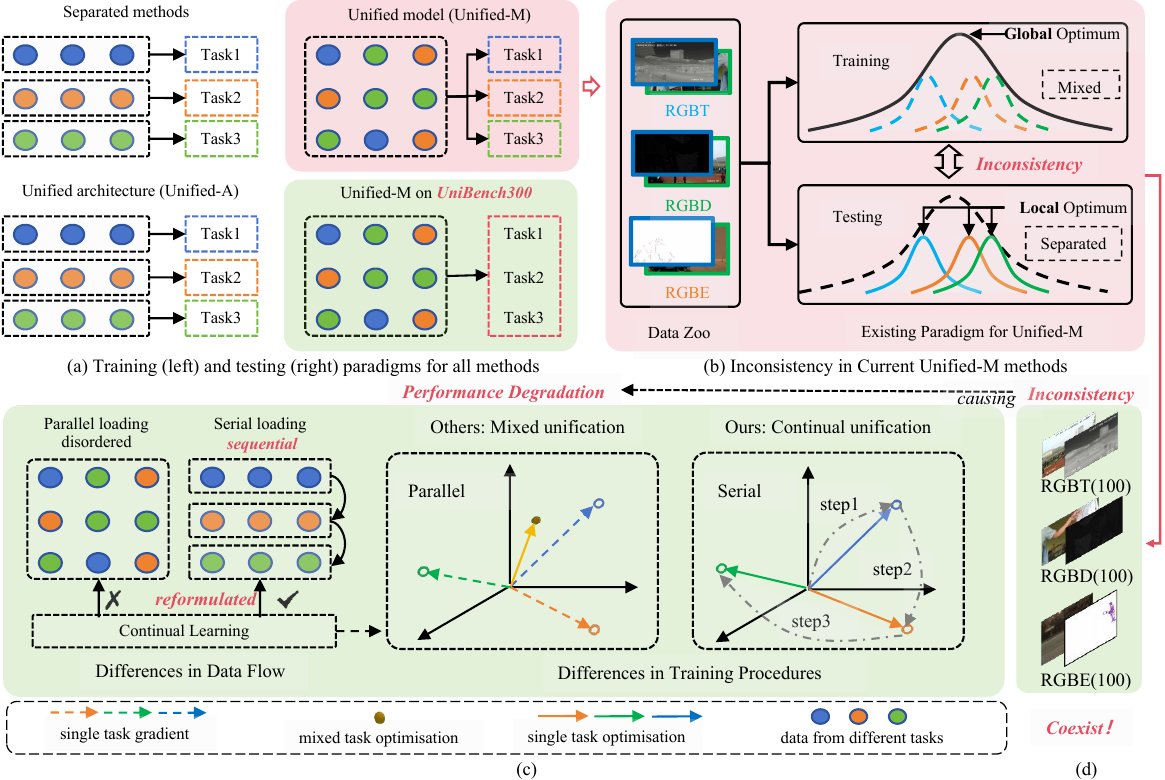}
  \caption{(a) Training (left) and testing (right) paradigms for all methods;
 (b) Inconsistency between the current training and testing paradigms (global vs. local) in unified methods (Unified-M) leads to performance degradation on separated benchmarks. To address these issues, (d) UniBench300 is proposed as the first unified benchmark to bridge the inconsistency, and (c) the unification process is reformulated as a serial one, thus facilitating the injection of CL to mitigate performance degradation. 
  }
  \label{fig:motivation}
\end{figure*}

Visual object tracking, which focuses on continuously predicting the object state (location and scale) throughout a video sequence, is advancing into the multi-modal era by leveraging the complementarity of various data modalities.
For example, compared with the well-known RGB modality, T modality enhances robustness against environmental changes \cite{lasher}, D modality provides 3D perception \cite{rgbd1k}, and E modality captures fine-grained motion cues \cite{visevent}.
Currently, there are three popular MMVOT tasks, RGBT, RGBD, and RGBE tracking, named after the involved modalities and have been demonstrated superior over RGB-only tracking \cite{moetrack, rgbd1k, visevent}.
This progress has sparked interest in unifying multiple MMVOT tasks to combine their strengths, aiming to improve robustness in challenging scenarios.

As shown in Figure~\ref{fig:motivation}(a), existing unified practices fall into two categories: methods with unified architectures (Unified-A) \cite{vipt, sdstrack, mixrgbx, protrack, emtrack, onetrack, minet} and those with unified models (Unified-M) \cite{unitrack, xtrack}.
Since Unified-A methods employ the same architectures across different tasks, they still require task-specific adaptations, leading to multiple independently trained models rather than a single unified model that integrates the advantages of all data modalities.
Therefore, further discussions primarily focus on the second category (Unified-M).

\textbf{Motivation:} Achieving a unified model starts with the most essential step—data preparation.
As shown in Figure~\ref{fig:motivation}(b), existing methods with unified models typically mix all types of data into a unified data zoo \cite{unitrack, xtrack}.
During training, different multi-modal data are loaded in parallel, infusing diverse multi-modal knowledge into the unified model to find a global optimum on the joint distribution.
However, this contradicts the testing phase, where unified models are evaluated on separated benchmarks, in the same way with methods that are trained separately \cite{qat, gmmt, bat, sttrack}.
It reflects that local optimum is preferred and this bias introduces an inconsistency between training and testing, ultimately leading to performance degradation, as demonstrated in Figure~\ref{fig:teaser} and Table~\ref{tab:comparison with unified trackers}.

To address these issues, our work makes two crucial advancements: 
\ding{182} As shown in Figure~\ref{fig:motivation}(d), to resolve the inconsistency issue, UniBench300 is introduced as the first unified benchmark for MMVOT.
It comprises 300 video sequences, including 100 RGBT sequences, 100 RGBD sequences, and 100 RGBE sequences, and 368.1K frames in total.
By forming a joint distribution of multi-modal data, UniBench300 aligns the training and testing paradigms, thus bridging the inconsistency.
Additionally, as quantified in Table~\ref{tab:platform}, UniBench300 provides convenient and efficient evaluation by reducing inference time by 27\% and requiring only a single evaluation pass (without performing three separate evaluations compared with previous formulations).
\ding{183} From a data-centric perspective, unification can be approached in either a parallel or serial manner.
While existing works \cite{unitrack, xtrack} adopt parallel unification, they suffer from performance degradation, necessitating the exploration of serial unification.
As depicted in Figure~\ref{fig:motivation}(c), serial unification progressively integrates new tasks, specifying performance degradation as knowledge forgetting of previous tasks, which is a core topic in the field of CL \cite{continuallearning2024}.
Based on this, reformulating the unification process in a serial one enables the natural incorporation of CL techniques into the unification of MMVOT tasks.  
As evidenced in Figure~\ref{fig:teaser}, results on two baselines and three benchmarks validate the superiority of CL in building a more stable unification process.

\textbf{Remark:} Notably, two intriguing phenomena emerge from our dedicated analyses:
\ding{172} Method 1 (ViPT) suffers more severe performance degradation than Method 2 (SymTrack) across all tasks; 
\ding{173} Both methods follow the same trend of degradation across tasks (RGBT > RGBD > RGBE).
For the first observation, further investigations in Table~\ref{tab:layers} suggest that the network capacity plays a critical role, where larger networks experiencing less degradation.
For the second one, qualitative analysis of data distributions in Figure~\ref{fig:modalities} and quantitative cross-task validation in Figure~\ref{fig:crossvalidation} reveal that D and E modalities are closer in distributions compared to T, and the discrepancy between D and T is larger than that between E and T.
This explains why RGBT tracking (combining two distant modalities, D and E) absorbs a more negative impact than RGBD and RGBE tasks (combining solely one relatively distant modality, T, and T is relatively closer to E than D).

In summary, the main contributions of our work include:

$\bullet$ UniBench300, the first unified benchmark integrating RGBT, RGBD, and RGBE data, is proposed to bridge the inconsistency between current training and testing paradigms of task unification. It offers a more convenient and efficient platform for method evaluation (inference once on UniBench300 with 93 minutes vs. three times on LasHeR, DepthTrack, and VisEvent with 127 minutes), significantly reducing the testing time by 27\%.

$\bullet$ Current parallel unification process is reformulated into a serial one from the data-centric perspective, based on which the integration of continual learning techniques can be naturally facilitated. This remarkably improves unification stability by mitigating knowledge forgetting of previous tasks.

$\bullet$  Two underlying reasons for performance degradation after unification are revealed, with potential suggestions for future research.: \ding{172} Performance degradation is negatively correlated to the network capacity. \ding{173} Modality discrepancies are responsible for the distinct degradation levels of diverse tasks. 

$\bullet$ Extensive experiments conducted on two baselines and four benchmarks demonstrate the significance of the proposed benchmark and the superiority of CL in stabilising the unification of MMVOT tasks. 

\section{Related Work}
\subsection{Multi-Modal Tracking Benchmark}
After recognising the advantages of multi-modal data in enhancing tracking performance under adverse scenarios, various benchmarks are introduced to advance MMVOT research, such as RGBT234 \cite{rgbt234}, LasHeR \cite{lasher}, RGBD1K \cite{rgbd1k}, DepthTrack \cite{depthtrack}, VisEvent \cite{visevent}, and COESOT \cite{coesot}.
While these benchmarks have significantly deepened research on MMVOT tasks, all of them focus on a single type of multi-modal data.
For instance, LasHeR, DepthTrack, and VisEvent solely contain RGBT, RGBD, and RGBE data, respectively.
However, this limitation hinders the development of unified trackers, which require a benchmark encompassing multiple MMVOT tasks.
As a result, existing unified methods are trained on mixed data but evaluated separately on each individual task, resulting in an inconsistency between training and testing.
To address this issue, we introduce UniBench300, the first benchmark that integrates RGBT, RGBD, and RGBE data, bridging this gap and facilitating further research on unified tracking.

\subsection{Unified Multi-Modal Visual Object Tracking}
Realising the complexity of practical scenarios, current studies are striving to unite the merits of multiple MMOVT tasks by designing unified trackers.
Existing approaches can be categorised into two groups based on training paradigms: (1) Unified-A: methods in this category keep the network architecture the same yet the parameters retrained when implementing on different tasks \cite{vipt, protrack, sdstrack, onetrack, emtrack, mixrgbx, minet}. 
As a result, they produce multiple task-specific models rather than a truly unified model. 
This limits their ability to handle diverse challenges, leaving them with the same limitations as separately trained methods, as shown in Figure~\ref{fig:motivation}(a).
(2) Unified-M: These methods adopt a single model capable of supporting multiple tasks \cite{unitrack, xtrack}, making them more suitable for deployment.
However, they directly mix all data types, which are then loaded in parallel during training without considering the uniqueness across modalities.
This leads to performance degradation compared with separately trained models.
To address this, we revisit the unification process from a data perspective and propose a serial unification paradigm that progressively integrates knowledge from new tasks, mitigating performance degradation by borrowing techniques from CL.

\subsection{CL in Multi-Modal Tracking}
CL focuses on the stable integration of knowledge from diverse tasks \cite{continuallearning2024, continuallearning2021} and has been successfully applied in various domains, such as multi-modal image fusion \cite{u2fusion, liu2024promptfusion, liu2025dcevo}, large-language models \cite{continualllm}, and single-modal visual object tracking \cite{continualtracking1, continualtracking2, odtrack, xu2025less}.
However, its application in multi-modal tracking, particularly in the unification of MMVOT tasks, remains under-explored.
As depicted in Figure~\ref{fig:motivation}(c), the primary limitation lies in the current training paradigm, where multi-modal data from different tasks are disorderly and parallelly utilised, failing to form a structured multi-task scheme, which is a prerequisite of applying CL.
In contrast, our work reformulates the unification process into a serial one, where task data appear progressively. 
\textit{This transformation bridges the gap between MMVOT unification and CL, paving the way for further exploration of CL techniques to enhance the stability of constructing unified models.}

\begin{table}[tb]
  \caption{Existing MMVOT benchmarks and UniBench300.
  }
  \label{tab:benchmarks}
  \centering
  \small
  \resizebox{\linewidth}{!}{
  \begin{tabular}{cccc}
    \toprule
    \multirow{2}{*}{\textbf{Benchmark}} & Sequence & Total & \multirow{2}{*}{X Modality} \\
    & Number   & Frames & \\
    \midrule
    GTOT \cite{gtot} & 50  & 7.8 K & T(100\%)  \\
    RGBT234 \cite{rgbt234} & 234  & 116.7 K & T(100\%)   \\
    LasHeR-test \cite{lasher} & 245  & 220.7 K & T(100\%)   \\
    VTUAV-ST-test \cite{hmft} & 176  & 631.4 K & T(100\%)    \\
    MV-RGBT \cite{moetrack} & 124  & 89.9 K & T(100\%)    \\
    CDTB \cite{cdtb} & 80  & 102.0 k & D(100\%)   \\
    DepthTrack-test \cite{depthtrack} & 50  &76.4 K  & D(100\%)   \\
    RGBD1K-test \cite{rgbd1k} & 50 & 117.9 K & D(100\%)   \\
    ARKittrack-test \cite{arkittrack} & 50 & 64.3 K & D(100\%)    \\
    FE108-test \cite{fe108} & 32 & 59.7 K & E(100\%)   \\
    VisEvent-test \cite{visevent} & 320 & 106.8 K & E(100\%)   \\
    COESOT-test \cite{coesot} & 528 & 176.6 K & E(100\%)   \\
    \midrule
\rowcolor[gray]{0.9}
    Ours(UniBench300) & 300 & 368.1K & T(33.3\%)/D(33.3\%)/E(33.3\%) \\
  \bottomrule
  
  \end{tabular}
  }
\end{table}

\section{UniBench300}
Data Collection: Recognising the inconsistency between the training and testing paradigms of current unified trackers \cite{unitrack, xtrack}, UniBench300 is proposed to bridge this gap by supporting multiple tracking tasks simultaneously.
Unlike existing benchmarks, as shown in Table~\ref{tab:benchmarks}, UniBench300 is the first benchmark to incorporate RGBT, RGBD, and RGBE sequences.
It consists of 300 sequences in total, including 100 RGBT sequences from LasHeR \cite{lasher}, 100 RGBE sequences from VisEvent \cite{visevent}, 50 RGBD sequences from DepthTrack \cite{depthtrack}, and 50 RGBD sequences from RGBD1K \cite{rgbd1k}.
Importantly, the sequence number across tasks is balanced to prevent potential bias.

As to the principle for data collection, we adhere to the principle of constructing a challenging benchmark that better drives progress in the MMVOT community. 
With this in mind, taking RGBT sequences for an example, we first rank all the sequences in LasHeR according to the performance (mean Interaction over the Union (IoU) of each sequence) of several advanced methods, including TBSI \cite{tbsi}, GMMT \cite{gmmt}, and BAT \cite{bat}.
The 100 sequences with the lowest performance are then selected as the RGBT split of UniBench300.
The RGBE split is obtained similarly from VisEvent, using TENet \cite{tenet} and SeqTrackV2 \cite{seqtrackv2} as representative RGBE trackers.
Finally, due to the limited size of DepthTrack and RGBD1K (both containing only 50 sequences in their testing sets), they are merged to form the RGBD split.
\textit{In summary, UniBench300 serves as a challenging split of existing RGBT, RGBD, and RGBE benchmarks yet with all modalities contained simultaneously.
This unique composition grants UniBench300 significant advantages over other benchmarks, offering improved convenience and efficiency for evaluation, as detailed in Table~\ref{tab:platform}.}

Evaluation Metrics: UniBench300 adopts precision rate (PR) and success rate (SR) as evaluation metrics, aligning with those used in established benchmarks such as VisEvent \cite{rgbt234} and LasHeR \cite{lasher}.
Thus, the detailed introduction of PR and SR are remained in the supplementary material.

\section{Methodology}
Unifying the merits of multiple data modalities is a promising approach to tackling high-dynamic challenges in real-world applications.
However, existing methods indiscriminately mix all data, neglecting their unique characteristics and leading to performance degradation.
In this work, we revisit the unification process from the perspective of data mixture distribution and reformulate a serial solution against the existing parallel paradigm.
This transformation allows performance degradation to be further interpreted as knowledge forgetting of previous tasks, which aligns with the philosophy of CL.
Consequently, experiments on two baselines validate the effectiveness of CL in promoting a more stable unification process.

\subsection{Unified Multi-Modal Tracker}
Following the widely used paradigm \cite{unitrack}, the training process for unified multi-modal trackers is formulated as minimising the distance between predicted and ground truth bounding boxes:
\begin{equation}
    \rm arg\min_{\theta} Loss(\textit{f}(\rm mix(d_{\textit{1}}, d_{\textit{2}},..., d_{\textit{n}}), \theta), \textbf{\textit{g}})
\end{equation}
where $f$ and $\theta$ are the tracker and its parameters, respectively.
$\rm d_{\textit{n}}$ denotes the training data from the $n^{th}$ task and \textbf{\textit{g}} represents ground truth bounding boxes for the involved data batch. 
$\rm mix(d_{\textit{1}}, d_{\textit{2}},..., d_{\textit{n}}$) refers to a disordered mixture of data from all tasks, which are then parallelly loaded for training, aiming to find a global optimum over the joint distribution of all tasks.
However, after training, the model is evaluated $n$ times on different tasks with each time expecting to reach a local optimum on a specific benchmark.
This reveals an inherent inconsistency between the training and testing paradigms, ultimately leading to the degraded performance of unified methods compared with those trained separately for individual tasks. 
As to the loss function, please refer to ViPT \cite{vipt} for more details.

\subsection{Continual Unification}
The performance degradation induced by the inconsistency between training and testing paradigms motivates us to delve into the unification procedure.
Basically, from a data-centric perspective, unification can be approached in two ways: parallel and serial.
Current practices \cite{unitrack, xtrack} follow a parallel paradigm by disorderly mixing and loading data, which leads to performance degradation.
Therefore, we reformulate the unification process by integrating knowledge from multiple tasks in a serial manner.
Under this paradigm, performance degradation can be further interpreted as knowledge forgetting of previous tasks which is a fundamental challenge in CL \cite{continuallearning2021}.
This naturally guides us to explore CL techniques to enhance the unification process, which is termed the continual unification process, as illustrated in Figure~\ref{fig:motivation}(c).
Specifically, we employ replay, a widely adopted CL technique \cite{continuallearning2024}, with its mathematical formulation given as follows:
\begin{equation}
 \rm arg\min_{\theta_{\textit{i}}} Loss(\textit{f}(\rm mix(d_{\textit{1}},...,d_{\textit{i}}), \theta_{\textit{i}-1}), \textbf{\textit{g}}), \textit{i} \in [1,...,\textit{n}] 
\label{eq:continual unification}
\end{equation}
where $\theta_{i-1}$ and $\theta_i$ are parameters learned from the ${(i-1)}^{th}$ and $i^{th}$ training steps, respectively.
In the beginning, at $i=1$, $\theta_\textit{0}$ is randomly initialised and only the training data ($d_\textit{1}$) from task 1 is utilised, which is the same as training a specific model for task 1. 
As a new task arrives, the previously trained model $\theta_{\textit{i-1}}$ is retained as an initialiser in the coming training step and the training data from previous tasks are also preserved.
This ensures that task unification is achieved progressively in a serial manner.
Moreover, by integrating CL, the final model $f(\theta_n)$ maintains performance comparable to separately trained models, significantly outperforming methods trained in a parallel fashion, as illustrated in Figure~\ref{fig:teaser}.

\textit{Discussions}: Eq~\ref{eq:continual unification} indicates that both the pre-trained model and data from previous tasks are replayed in the continual unification process. The rationale behind this is twofold: \ding{172} The pre-trained model must be retained; otherwise, when $i=n$, the continual unification process is the same as the original parallel paradigm, which suffers from performance degradation; \ding{173} The data from previous tasks should be incorporated; otherwise, the model will suffer from catastrophic forgetting, as highlighted in studies of CL \cite{continuallearning2021, continuallearning2024}.

\subsection{Differences in Training Paradigms}
Figure~\ref{fig:motivation} intuitively illustrates the differences between our continual unification and the widely used paradigm.
In the existing paradigm, all types of data are typically mixed within each batch, leading to multiple optimisation directions after gradient computation (mixed unification).
The gradients are then simply averaged based on batch size \cite{unitrack} or task-specific sample counts \cite{xtrack}, disregarding deeper task discrepancies such as the appropriate magnitude of each update per task. 
As a result, the final optimisation (line in yellow) is suboptimal, leading to insufficient unification.
In contrast, the proposed continual unification process adopts a multi-step training strategy.
When a new task is introduced, the CL technique is activated, enabling the model to learn new knowledge while better-preserving performance on previously learned tasks.
\textit{To summarise, these two paradigms differ in one key aspect: From a data-centric perspective, the existing paradigm loads training data in a parallel manner, whereas the proposed approach processes it sequentially.}

Figure~\ref{fig:motivation}(c) exhibits the differences in data flow. 
\textit{It is evident that merely by reformulating the unification process into a serial paradigm, it aligns with the typical scenarios (multi-task) studied in the realm of CL \cite{continuallearning2024}.}
The significant improvements achieved through CL, as shown in Figure~\ref{fig:teaser}, further validate the correctness of revisiting the unification process from a data-centric perspective.

\begin{figure}[t]
  \centering
  \includegraphics[width=0.9\linewidth]{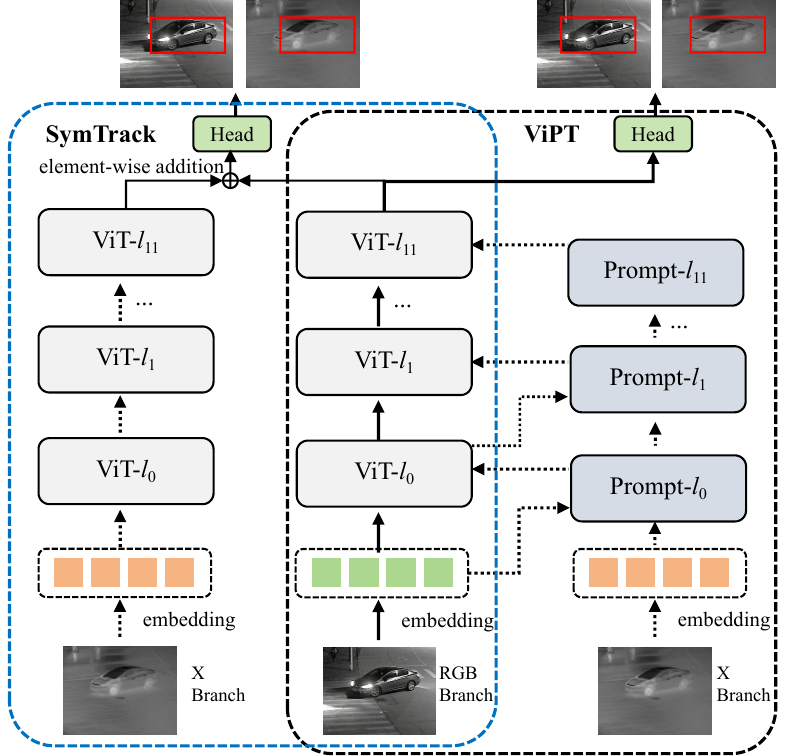}
  \caption{Architectures of SymTrack and ViPT, that share a standard RGB branch, as both methods process RGB data in the same manner.}
  \label{fig:methods}
\end{figure}

\subsection{Implemented Baselines}
As shown in Figure~\ref{fig:methods}, the proposed continual unification for MMVOT tasks is implemented on two baselines: ViPT \cite{vipt}, a well-known asymmetric method adopted as the baseline of many unified trackers \cite{xtrack, unitrack, onetrack}, and SymTrack, a self-designed baseline named after its symmetric structure.
Both methods follow the standard "embedding-backbone-head" pipeline, where input images are first transformed into an embedding space, processed by a backbone, and then projected into the task space through a tracking head.
While ViPT and SymTrack share the same processing strategy for RGB data, they differ in handling X data ( X can be T, D, or E).
Specifically, both methods employ a pre-trained ViT-B \cite{ostrack} with twelve self-attention layers for the RGB branch.
For the X branch, ViPT incorporates twelve lightweight prompt layers with each of them consisting of three convolutional layers.
In contrast, SymTrack adopts a symmetric architecture where RGB and X branches share parameters.
Additionally, in ViPT, only the output of the RGB branch is fed into the tracking head, whereas in SymTrack, outputs of both RGB and X branches are utilised.

\begin{table*}[t]
  \centering
  \caption{Comparisons with multi-modal trackers on VisEvent, DepthTrack, and LasHeR benchmarks. "Separated" denotes that the method is tailored for a specific task. Superscript "*" represents that the method is reproduced in this study. 
  }
  \resizebox{1\linewidth}{!}{
\begin{tabular}{ccccccccccccc}
\toprule
\multicolumn{1}{c}{\multirow{2}{*}{Type}}&\multicolumn{1}{c}{\multirow{2}{*}{Method}} &\multicolumn{1}{c}{\multirow{2}{*}{Venue}}& \multicolumn{2}{c}{\multirow{1}{*}{Prior}} & \multicolumn{3}{l}{LasHeR}  & \multicolumn{3}{l}{DepthTrack} & \multicolumn{2}{l}{VisEvent} \\ 
&\multicolumn{1}{c}{}& & Train& Test & PR $\uparrow$& NPR $\uparrow$& SR $\uparrow$& Pr $\uparrow$ & Re $\uparrow$& F-score $\uparrow$ & PR $\uparrow$ & SR $\uparrow$\\
\midrule
\multicolumn{1}{c}{\multirow{6}{*}{Separated}} &GMMT \cite{gmmt}&AAAI'2024& \usym{1F5F8} & \usym{1F5F8}&0.707&0.670& 0.566 & -& - & - &-& -\\
&BAT \cite{bat}&AAAI'2024& \usym{1F5F8} & \usym{1F5F8}&0.702&-&0.563 &-&- &- &-& -\\
&SSLTrack \cite{ssltrack}&PR'2024& \usym{1F5F8} & \usym{1F5F8}&-&-&- & 0.565& 0.491 & 0.525 &-& -\\
&VADT \cite{vadt}&ICASSP'2024& \usym{1F5F8} & \usym{1F5F8}&-&-&- & 0.606& 0.603 & 0.610 &-& -\\
&eMoE-Tracker \cite{emoe-track}&Arxiv'2024& \usym{1F5F8} & \usym{1F5F8}&-&-&- & -& - & - &0.764& 0.613\\
&TENet \cite{tenet}&NN'2025& \usym{1F5F8} & \usym{1F5F8}&-&-&- & -& - & - &0.765& 0.601\\
\midrule
\multicolumn{1}{c}{\multirow{7}{*}{Unified-A}} &ProTrack \cite{protrack}&ACMMM'2022& \usym{1F5F8} & \usym{1F5F8}&0.509&-& 0.421 &0.583&0.573&0.578&0.617& 0.474\\
&ViPT* \cite{vipt}&CVPR'2023& \usym{1F5F8} & \usym{1F5F8}&0.645&0.614& 0.519 & 0.587&0.611 &0.598&0.754& 0.591\\
&MixRGBX \cite{mixrgbx} & NEUCOM'2024 &\usym{1F5F8} & \usym{1F5F8}&0.672& - & 0.536&0.593 & 0.609 &0.601&0.774& 0.602\\
&EMTrack \cite{emtrack} & TCSVT'2024 &\usym{1F5F8} & \usym{1F5F8}&0.659& - & 0.533&0.580 & 0.585 &0.583&0.724& 0.584\\
&SDSTrack \cite{sdstrack}&CVPR'2024& \usym{1F5F8} & \usym{1F5F8}&0.665&0.631& 0.531 &0.619&0.609&0.614&0.767& 0.597\\
&OneTracker \cite{onetrack} & CVPR'2024 &\usym{1F5F8} & \usym{1F5F8}&0.672& - & 0.538&0.607 & 0.604 &0.609&0.767& 0.608\\
&SymTrack & - &\usym{1F5F8} & \usym{1F5F8}&0.708& 0.670 & 0.569&0.592&0.617&0.604&0.770& 0.606\\
\midrule
\multicolumn{1}{c}{\multirow{7}{*}{Unified-M}}&Un-Track \cite{unitrack} & CVPR'2024 &\usym{1F5F8} & \usym{1F5F8}&0.646& - & 0.513& 0.610 & 0.610 &0.610&0.755& 0.589\\
&XTrack \cite{xtrack} & Arxiv'2024 &\usym{1F5F8} & \usym{2717}&0.655& - & 0.525& 0.597 & 0.597 &0.598&0.756& 0.591\\
&ViPT*+mixed&CVPR'2023& \usym{2717} & \usym{2717}&0.609&0.576& 0.494 & 0.562& 0.584 & 0.573 &0.743& 0.579\\
&ViPT*+CL &-& \usym{1F5F8} & \usym{2717}&0.652&0.618& 0.527 & 0.584& 0.608 & 0.596 &0.758& 0.592\\
\multicolumn{5}{c}{ }  &\textbf{+4.3\%}&\textbf{+4.1\%} & \textbf{+3.3\%}& \textbf{+2.2\%} &\textbf{+2.4\%} &\textbf{+2.3\%}&\textbf{+1.5\%}& \textbf{+1.3\%}\\
&SymTrack+mixed & - &\usym{2717} & \usym{2717}&0.682& 0.649 & 0.550&0.581&0.605&0.593&0.763& 0.597\\
&SymTrack+CL & - &\usym{1F5F8} & \usym{2717}&0.714& 0.676 & 0.573&0.587 &0.613&0.600&0.771&0.607 \\
\multicolumn{5}{c}{ } &\textbf{+3.2\%}&\textbf{+2.7\%} & \textbf{+2.3\%}& \textbf{+0.6\%} &\textbf{+0.8\%} &\textbf{+0.7\%}&\textbf{+0.8\%}& \textbf{+1.0\%}\\
\bottomrule
\end{tabular}
} 
\label{tab:comparison with unified trackers}
\end{table*}

\begin{table*}[t]
  \centering
  \caption{Observations after transferring ViPT* and SymTrack from methods with unified architectures to unified models.
  }
  \resizebox{1\linewidth}{!}{
\begin{tabular}{cccccccccccccc}
\toprule
\multicolumn{1}{c}{\multirow{2}{*}{Type}}&\multicolumn{1}{c}{\multirow{2}{*}{Method}} & \multicolumn{3}{l}{LasHeR} & \multicolumn{1}{c}{\multirow{2}{*}{Mean}}& \multicolumn{3}{l}{DepthTrack} & \multicolumn{1}{c}{\multirow{2}{*}{Mean}}&\multicolumn{2}{l}{VisEvent} &\multicolumn{1}{c}{\multirow{2}{*}{Mean}}&\\ 
&\multicolumn{1}{c}{}& PR $\uparrow$& NPR $\uparrow$& SR $\uparrow$&& Pr $\uparrow$ & Re $\uparrow$& F-score $\uparrow$ && PR $\uparrow$ & SR $\uparrow$&&\\
\midrule
Unified-A &ViPT*&0.645&0.614& 0.519 && 0.587&0.611 &0.598&&0.754& 0.591&&\multicolumn{1}{c}{\multirow{2}{*}{\textbf{Degradation:}}}\\
Unified-M &ViPT*+mixed &0.609&0.576& 0.494 & &0.562& 0.584 & 0.573 &&0.743& 0.579&\\
& &\textbf{-3.6\%}&\textbf{-3.8\%} & \textbf{-2.5\%}&\textbf{-3.30\%}& \textbf{-2.5\%} &\textbf{-2.7\%} &\textbf{-2.5\%}&\textbf{-2.57\%}&\textbf{-1.1\%}& \textbf{-1.3\%}&\textbf{-1.20\%}&\multicolumn{1}{c}{\multirow{1}{*}{\textbf{ViPT*}}}\\
Unified-A &SymTrack &0.708& 0.670 & 0.569&&0.592&0.617&0.604&&0.770& 0.606&&\multicolumn{1}{c}{\multirow{1}{*}{\textbf{$\vee$}}}\\
Unified-M &SymTrack+mixed &0.682& 0.649 & 0.550&&0.581&0.605&0.593&&0.763& 0.597&&\multicolumn{1}{c}{\multirow{1}{*}{\textbf{SymTrack}}}\\
& &\textbf{-2.6\%}&\textbf{-2.1\%} & \textbf{-1.9\%}&\textbf{-2.20\%}& \textbf{-1.1\%} &\textbf{-1.2\%} &\textbf{-1.1\%}&\textbf{-1.13\%}&\textbf{-0.7\%}& \textbf{-0.9\%}&\textbf{-0.80\%}&\\
\midrule
&&\multicolumn{11}{c}{\multirow{1}{*}{\textbf{Degradation: RGBT > RGBD > RGBE}}}&\\
\bottomrule
\end{tabular}
} 
\label{tab:observations}
\end{table*}

\section{Experiments} \label{sec:experiment}
\subsection{Implementation Details}
\textit{Training Details:} This work is implemented on a platform with an NVIDIA RTX 3090 GPU and an AMD R9 7950X CPU.
Specifically, during training, the epoch and batch size are set to 60 and 32, respectively.
As to other hyperparameters, for ViPT \cite{vipt}, we follow its official implementation, while for SymTrack, we retain the same configurations as MPLT \cite{mplt}.
\textit{Datasets:} To ensure a fair comparison, our training sets are kept the same with other unified trackers, including VisEvent \cite{visevent}, LasHeR \cite{lasher}, and DepthTrack \cite{depthtrack}.
These datasets are used separately for training "Unified-A" variants and simultaneously for the "Unified-M" category.
Later, the evaluation is conducted on the test splits of these datasets as well as the proposed UniBench300.
\textit{Evaluation Metrics:}
The well-known PR \cite{gtot}, normalised precision rate (NPR) \cite{trackingnet}, and SR \cite{gtot} are employed for evaluations on LasHeR, VisEvent and UniBench300, and precision (Pr) \cite{rgbd1k}, recall (Re) \cite{rgbd1k}, and F-score \cite{rgbd1k} are utilised for DepthTrack.
In general, PR, NPR, Pr, and Re measure the accuracy of the predictions.
SR evaluates the ratio of successfully tracked frames.
F-score combines both precision (Pr) and recall (Re), providing a comprehensive indicator for ranking.

\subsection{Significance of UniBench300}
\begin{figure}[t]
  \centering
  \includegraphics[width=1.0\linewidth]{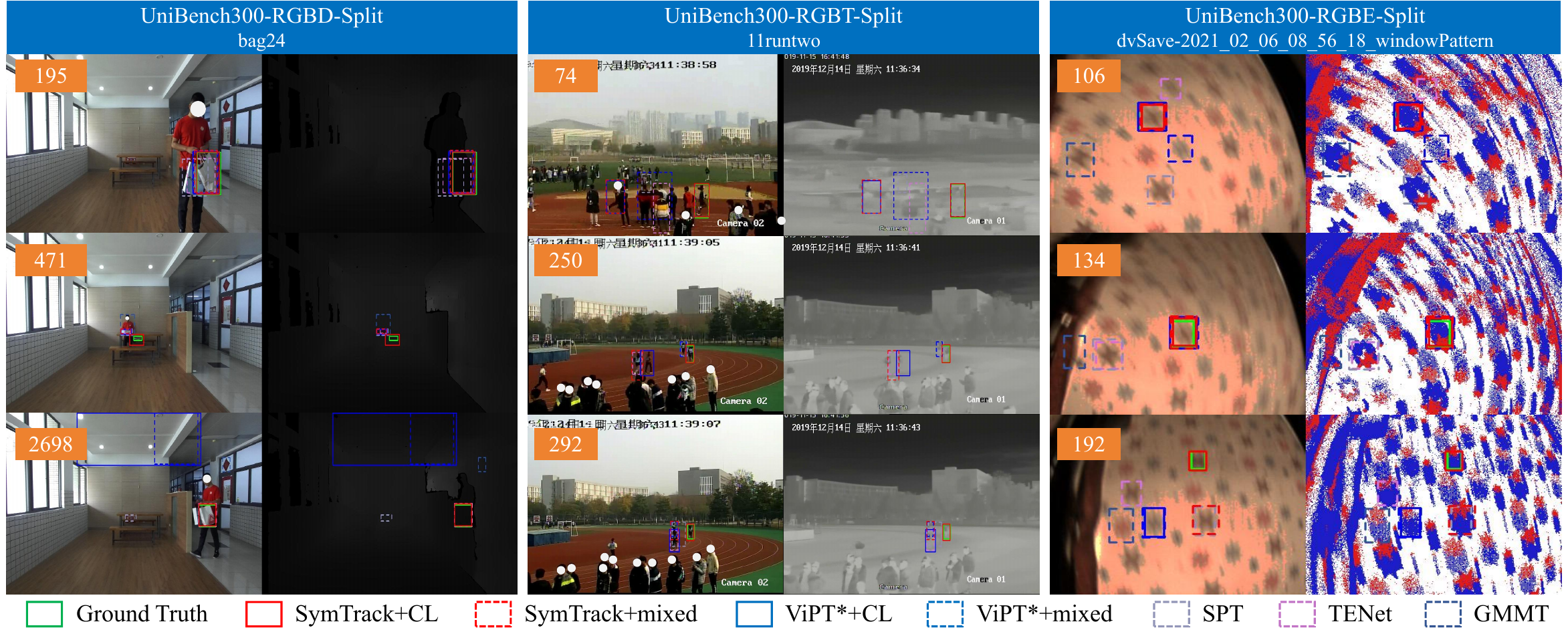}
  \caption{Visualisations on UniBench300.}
  \label{fig:vis}
\end{figure}

\textit{Unified Benchmark}: Table~\ref{tab:benchmarks} significantly highlights the advantages of UniBench300 as the first unified benchmark that incorporates multi-task data.
This enables seamless evaluation of both separated and unified methods on this benchmark.
Figure~\ref{fig:vis} illustrates the tracking results of advanced separated and unified methods, where methods with unified models evidently outperform separated methods and methods with unified architectures, emphasising the importance of developing unified models for better robustness.

\textit{Convenient and Efficient}: Table~\ref{tab:platform} compares the evaluation time of ViPT* and SymTrack on existing benchmarks and UniBench300.
For example, ViPT* requires three separate evaluations on separate benchmarks such as LasHeR, DepthTrack, and VisEvent, costing 65, 24, and 38 minutes, respectively (totally 127 minutes). 
In contrast, it is executed only once on UniBench300, which consumes 93 minutes, resulting in a time savings of 26.77\% for inference.
Similarly, SymTrack shows a 27.03\% reduction in time consumption when evaluated on UniBench300, intuitively underscoring the convenience and efficiency induced by the proposal of UniBench300.

\textit{Challenging}: 
Figure~\ref{fig:res-unibench300} showcases the benchmarking results on UniBench300.
It is evident that results on UniBench300 are significantly lower than those on other benchmarks in Table~\ref{tab:comparison with unified trackers} (0.395 vs. 0.570, 30\% lower), facilitating a more challenging benchmark with greater potential to advance the MMVOT community.

\begin{table}[t]
  \caption{Comparisons of time consumption.}
  \centering \label{tab:platform}
  \resizebox{1\linewidth}{!}{
\begin{tabular}{c|ccc|c|c}
\toprule
Method & LasHeR & DepthTrack&VisEvent &UniBench300&$\Delta$ \\
\midrule
ViPT*&65min&24min&38min&93min&-26.77\%\\
SymTrack&76min&28min&44min&108min&-27.03\%\\
\bottomrule
\end{tabular}
}
\end{table}

\subsection{Superiority of Continual Unification}\label{sec:continualunification}

\begin{figure}[t]
  \centering
  \includegraphics[width=1.0\linewidth]{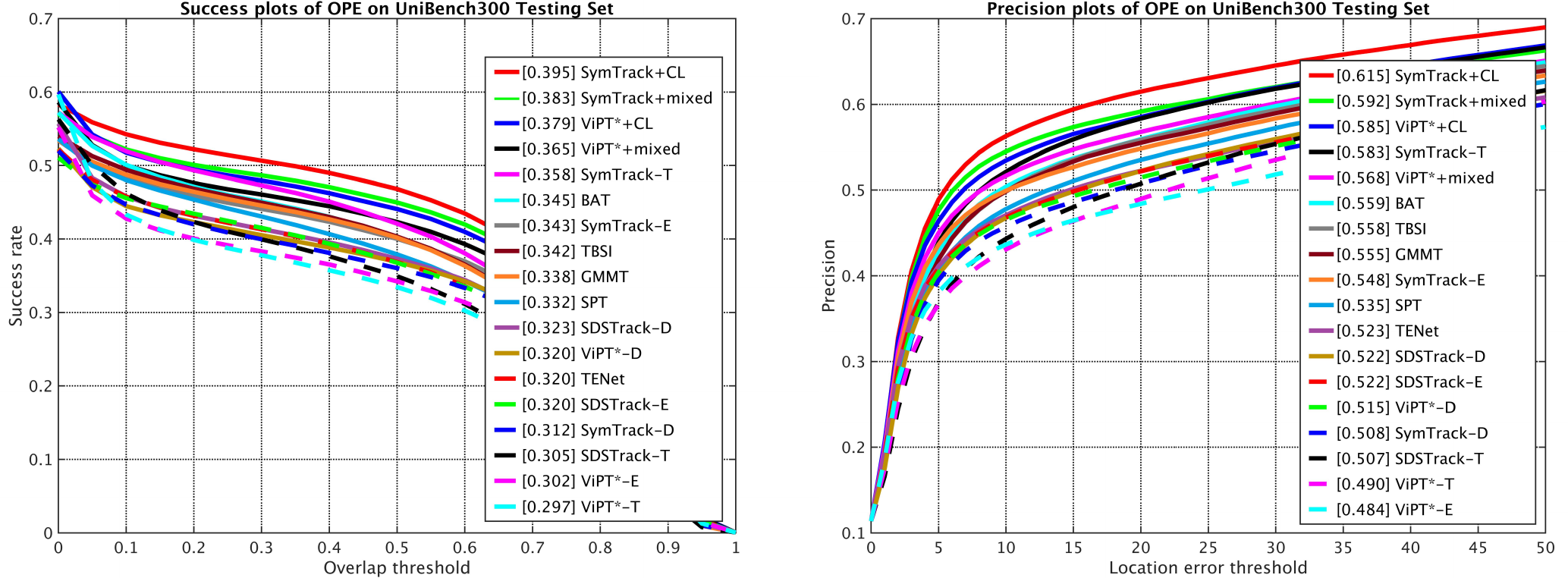}
  \caption{Benchmarking results on UniBench300.}
  \label{fig:res-unibench300}
\end{figure}

\textit{Advanced performance}: Table~\ref{tab:comparison with unified trackers} compares the performance of separated methods, methods with unified architectures and models across multiple benchmarks, including LasHeR \cite{lasher}, DepthTrack \cite{depthtrack}, and VisEvent \cite{visevent}, where "Prior" indicates notification of the task type.
It reveals that separated methods or those in the "Unified-A" category generally outperform methods in the "Unified-M" category. 
This difference can be attributed to the failure to account for modality discrepancies during unification.
Further details of performance degradation are presented in Table~\ref{tab:observations}.
Both ViPT* and SymTrack show reduced efficacy across all involved tasks. 
However, when trained under a serial paradigm, their performance improves, becoming comparable to or even better than their previous variants. 
Specifically, SymTrack+CL consistently outperforms SymTrack+mixed by 2.3\%, 0.7\%, and 1.0\% on LasHeR, DepthTrack, and VisEvent, respectively. 
Notably, on LasHeR, SymTrack+CL outperforms SymTrack by 0.6\% on PR (0.714 vs. 0.708), 0.6\% on NPR (0.676 vs. 0.670), and 0.4\% on SR (0.573 vs. 0.569).
Additionally, Figure~\ref{fig:res-unibench300} presents benchmarking results on UniBench300, where the suffixes "T", "D", and "E" denote variants tailored to RGBT, RGBD, and RGBE tasks, respectively.
From this figure, it is evident that SymTrack+CL outperforms all the competitors, achieving scores of 0.395 on SR and 0.615 on PR. 
It surpasses the runner-up (SymTrack+mixed) by 1.2\% on SR and 2.3\% on PR, further highlighting the effectiveness of CL in stabilising the unification process.

\textit{Fine-grained analysis:} In contrast to existing methods that load multi-task data in parallel, our continual unification process progressively integrates knowledge from new tasks.
This approach allows for a fine-grained analysis, as shown in Figure~\ref{fig:teaser}.
Unlike other methods that evaluate variants trained for one or three tasks (a total of four: T, D, E, and mix(T, D, E)), we experiment with all possible permutations and combinations within the continual unification framework. 
This comprehensive approach significantly demonstrates the superiority of CL in unifying MMVOT tasks by enhancing the performance of all fifteen variants.


\subsection{Insights for Unification}
From Table~\ref{tab:comparison with unified trackers}, it is evident that CL significantly benefits in recovering the performance of ViPT*+mixed and SymTrack+mixed.
Notably, two phenomena are observed: SymTrack recovers less than ViPT*, and their performance improvements on LasHeR are larger more than those on DepthTrack and VisEvent. 
This insight motivates a further exploration of the unification process.
To be more intuitive, Table~\ref{tab:observations} presents the performance degradation from ViPT* and SymTrack to ViPT*+mixed and SymTrack+mixed.
This table shows that ViPT* (SymTrack) experiences drops of 3.30\% (2.20\%), 2.57\% (1.13\%), and 1.2\% (0.80\%) on LasHeR, DepthTrack, and VisEvent, respectively. 
It suggests that, after unification, ViPT* suffers larger performance drops than SymTrack (ViPT* > SymTrack). 
Additionally, both methods show the greatest performance drop on LasHeR, followed by DepthTrack, and the smallest on VisEvent (RGBT > RGBD > RGBE).

\textbf{Degradation:ViPT*>SymTrack}: Figure~\ref{fig:methods} illustrates the architectures of ViPT* and SymTrack.
Both share the same architecture for the RGB branch but differ in the X branch.
This leads to a reasonable assumption that the performance degradation is related to the size of the X branch because ViPT* utilises lightweight prompt layers with solely three convolutional layers while SymTrack employs transformer blocks with quadratic complexity.
It also conforms to the intuition that a larger network has greater ability to accommodate more knowledge.
Building on this, experiments are conducted with progressively reduced layers for the X branch, keeping the RGB branch unchanged, and the results are presented in Table~\ref{tab:layers}.
As the network depth decreases from 12 to 2 layers, the performance degradation gradually increases from 2.20\% to 3.00\%. 
This supports our assumption and leads to the conclusion that a larger network experience less performance degradation after unification. 
This also provides an intuitive explanation for why ViPT* experiences larger degradations compared with SymTrack and offers valuable insights for future studies on unification.

\begin{figure}[t]
  \centering
  \includegraphics[width=0.9\linewidth]{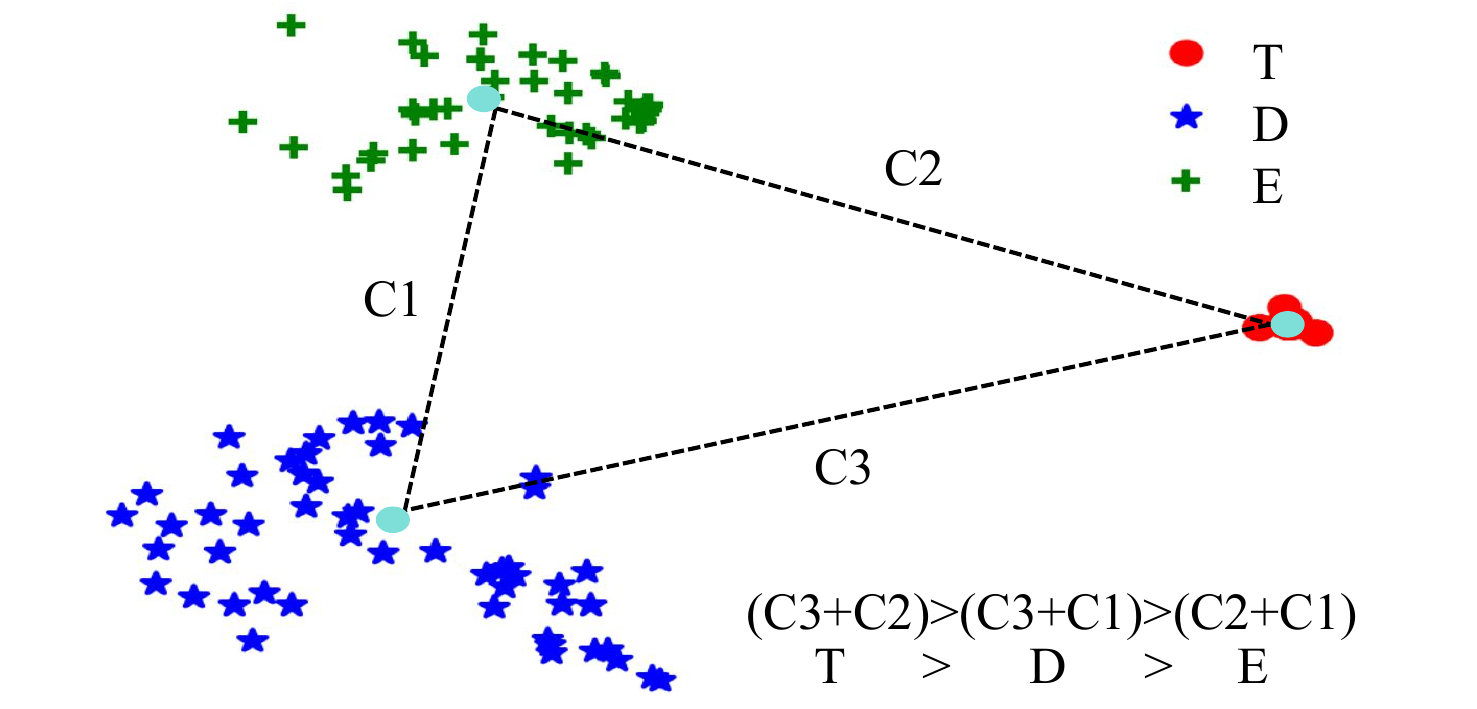}
  \caption{Distribution of T, D, and E data. C1, C2, and C3 represent the distance between modalities.}
  \label{fig:modalities}
\end{figure}

\textbf{Degradation:RGBT>RGBD>RGBE}: The most notable discrepancy among these tasks lies in the data type of the X branch.
Therefore, the differences in data modalities are likely the cause of this phenomenon.
Based on the results, the sum of the distances between T/D/E and the other two modalities should gradually decrease because larger distances indicate less similarity, which typically leads to worse performance in a multi-modal system \cite{zheng2023toward, aptrack, tmctrack, dfat, melt, }. 
Figure~\ref{fig:modalities} validates this by visualising the distributions of T, D, and E data.
To quantify the differences, C1, C2, and C3 are further introduced.
For example, C2 represents the distance between T and E, C3 measures the distance between T and D, and their combination (C3+C2) reflects the summarised distance of T to D and E.
From this figure, a clear pattern emerges: (C3+C2) > (C3+C1) > (C2+C1).
This means T, D, and E has sequentially decreased distance to another two modalities, which is consistent with the experimental observations.

\begin{figure}[t]
  \centering
  \includegraphics[width=0.9\linewidth]{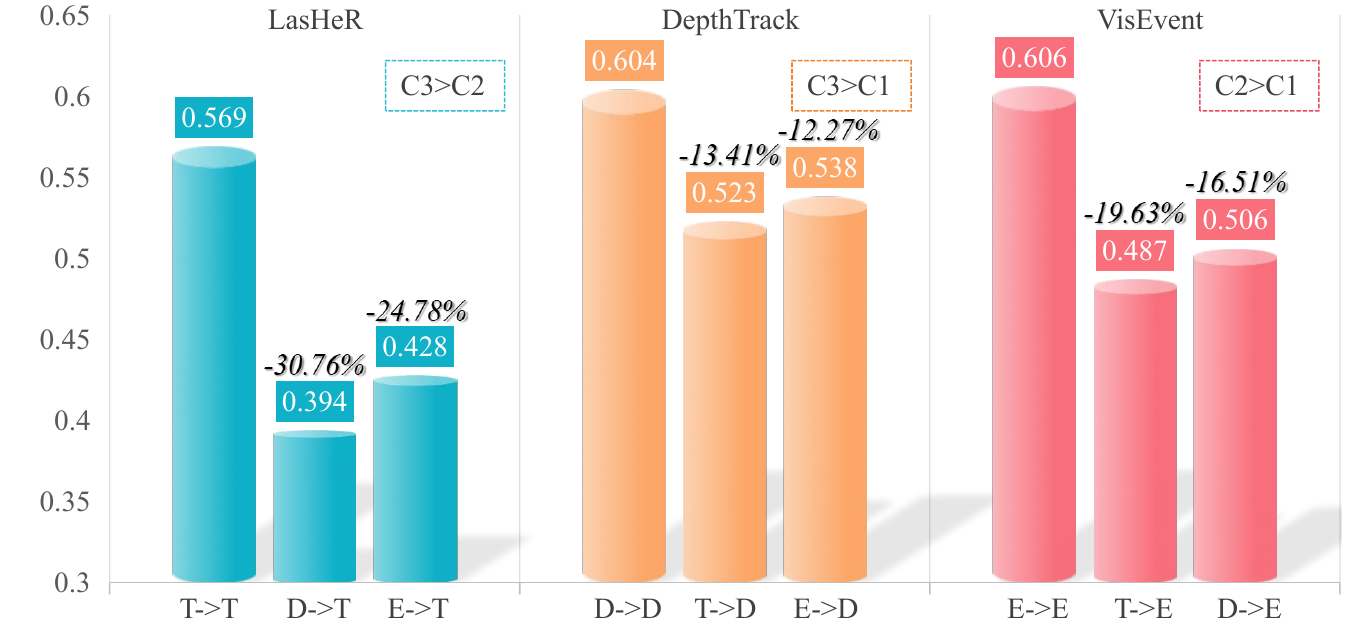}
  \caption{Cross validation of the modality discrepancy. D->T denotes applying the model trained with RGBD data to RGBT benchmark.}
  \label{fig:crossvalidation}
\end{figure}

Furthermore, to eliminate subjectivity in Figure~\ref{fig:modalities}, such as the data selected for visualisation or the choice of central points (light blue points) for distance calculation, additional cross-validation experiments are conducted.
The results are presented in Figure~\ref{fig:crossvalidation}.
On the left side, for LasHeR, the model trained with RGBT data achieves 0.569 on SR (T->T). 
Models trained with RGBD (D->T) and RGBE (E->T) data reach 0.394 and 0.428, respectively. 
Compared with T->T, the performance drops by 30.76\% for D->T and 24.78\% for E->T. 
This indicates that the discrepancy between T and D is larger than that between T and E (C3 > C2).
Analogously, the same analysis is applied to the middle and right parts, resulting in the conclusions (C3 > C1) and (C2 > C1).
Based on these, the cross-validation results confirm the conclusion (C3 + C2) > (C3 + C1) > (C2 + C1), which aligns with the findings from Figure~\ref{fig:modalities}, further validating the objectivity and reliability of our analysis.

\begin{table}[t]
  \caption{Degradation and network depth on LasHeR.}
  \centering \label{tab:layers}
  \resizebox{1\linewidth}{!}{
\begin{tabular}{cccccccccc}
\toprule
\multicolumn{1}{c}{\multirow{2}{*}{Layers}} & \multicolumn{3}{l}{LasHeR}  &{\multirow{2}{*}{Mean}}&\multicolumn{1}{c}{\multirow{2}{*}{Layers}} & \multicolumn{3}{l}{LasHeR}  &{\multirow{2}{*}{Mean}} \\
& PR $\uparrow$& NPR $\uparrow$& SR $\uparrow$&&& PR $\uparrow$& NPR $\uparrow$& SR $\uparrow$&\\
\midrule
\multicolumn{1}{c}{\multirow{3}{*}{12}}   & 0.708 & 0.670 & 0.569 &&\multicolumn{1}{c}{\multirow{3}{*}{6}}   & 0.690& 0.651 & 0.552 &\\
& 0.682& 0.649 & 0.550 &&& 0.658& 0.623 & 0.528 &\\
& -2.6\% & -2.1\% & -1.9\%&-2.20\% && -3.2\% & -2.8\% & -2.4\%&-2.80\%\\
\midrule
\multicolumn{1}{c}{\multirow{3}{*}{10}}   & 0.698& 0.659 & 0.557 & &\multicolumn{1}{c}{\multirow{3}{*}{4}}   & 0.685& 0.643 & 0.545 & \\
& 0.667& 0.637 & 0.537 &&&0.650& 0.615 & 0.522  &\\
& -3.1\% & -2.2\% & -2.0\%&-2.43\%&& -3.5\% & -2.8\% & -2.3\%&-2.87\%\\
\midrule
\multicolumn{1}{c}{\multirow{3}{*}{8}}   & 0.697& 0.657 & 0.557 &&\multicolumn{1}{c}{\multirow{3}{*}{2}}   & 0.673& 0.632 & 0.538 &\\
& 0.664& 0.631 & 0.535 &&& 0.637& 0.602 & 0.514 &\\
& -3.3\% & -2.6\% & -2.2\%&-2.70\%&& -3.6\% & -3.0\% & -2.4\%&-3.00\%\\
\bottomrule
\end{tabular}
}
\end{table}


In conclusion, our in-depth analyses demonstrate that performance degradation after unification is negatively correlated with network size and is task-related (tasks involving more heterogeneous data suffer greater degradations).

\subsection{What's in Supplementary Material}
Limited by the page, we leave the following information in the supplementary material:
\ding{172} Efficiency Analysis.
\ding{173} The results on all the previous tasks, which complements Figure~\ref{fig:teaser} (merely showing the performance of the primary task).
\ding{174} Qualitative analysis of the superiority of CL in the embedding space.
\ding{175} Pseudo code for the proposed continual unification procedure.
\ding{176} Evaluation metrics for UniBench300.
\ding{177} Insights for the sequence in continual unification.

\section{Conclusion}
In this work, we delve into the unification of MMVOT tasks and reveal the inconsistency between current training and testing paradigms, which leads to performance degradation. 
To address these issues, we introduce UniBench300, the first unified benchmark that simultaneously incorporates RGBT, RGBD, and RGBE data for multi-modal tracking.
This benchmark resolves the inconsistencies by aligning inference with the training process, which offers a more convenient and efficient evaluation platform that requires only a single inference (previously three times on separated benchmarks, including LasHeR, DepthTrack, and VisEvent), reducing time by 27\%.
Moreover, the original training paradigm is reformulated as a serial one from the data-centric perspective, which integrates new tasks progressively, framing performance degradation as a key challenge in CL: knowledge forgetting of previous tasks.
This naturally motivates the injection of CL into the unification process, resulting in significantly enhanced stability across all tasks involved.
Along with the extensive experiments, we identify two important insights: performance degradation after unification is negatively correlated with network size, and it is task-dependent, with tasks involving more heterogeneous data experiencing greater degradation, providing valuable suggestions for future research. Supplementary material can be found at https://github.com/Zhangyong-Tang/UniBench300/blob/main/Supplementary-material.pdf.

\begin{acks}
This work was supported in part by the National Key Research and Development Program of China (Grant No. 2023YFE0116300, 2023YFF1105102, 2023YFF1105105), the National Natural Science Foundation of China (Grant NO. 62020106012, 62332008, 62106089, 62336004), the 111 Project of Ministry of Education of China (Grant No.B12018), the Fundamental Research Funds for the Central Universities (JUSRP202504007), the Wuxi Science and Technology Development Fund Project (K20241025), and the Leverhulme Trust Emeritus Fellowship EM-2025-069\textbackslash9.
\end{acks}

\bibliographystyle{ACM-Reference-Format}
\balance
\bibliography{ref}


\end{document}